\documentclass[runningheads]{llncs}
\usepackage[T1]{fontenc}
\usepackage{graphicx,verbatim}
\usepackage{hyperref}       
\usepackage{url}            
\usepackage{booktabs}       
\usepackage{amsfonts}       
\usepackage{nicefrac}       
\usepackage{microtype}      
\usepackage{xcolor}         
\usepackage{amsmath}
\usepackage{multirow}
\usepackage{float}
\begin{document}
\title{\texorpdfstring{$K$}{K}-NeAS: Scalable Multi-Material CT Reconstruction Using Neural SDFs}
\author{Daksh K. Shah \and Emmanouil Nikolakakis \and Razvan Marinescu}
    \authorrunning{D.K. Shah et al.}
    \institute{University of California, Santa Cruz\\
    \email{daksh@ucsc.edu}\\
}

\maketitle              
\begin{abstract}
    Computed Tomography (CT) carries significant ionizing radiation risks, driving the need for sparse-view reconstruction. Implicit scene representations (ISRs) address this by recovering continuous volumetric attenuation fields directly from sparse projections, and recent geometry-aware extensions jointly model surface geometry alongside attenuation to improve fidelity and enable clean tissue segmentation without manual thresholding. However, these methods remain limited by manually tuned attenuation bounds and rigid two-material constraints. This paper proposes $K$-NeAS, a unified and scalable architecture for automated, multi-material surface reconstruction. We replace independent material networks with a shared latent backbone and introduce a fully differentiable $K$-material sequential soft selector to model an arbitrary number of overlapping tissues. To eliminate manual tuning, we automate attenuation bounding using a Gaussian Mixture Model (GMM) and implement a scheduled auxiliary floater loss to mitigate geometric hallucinations common under extreme sparsity. Evaluated across four simulated Cone-Beam CT (CBCT) datasets, $K$-NeAS successfully scales to arbitrary material counts, achieving superior 3D volumetric fidelity at $K=3$ materials on complex multi-tissue regions such as the Abdomen ($33.28\text{ dB}$ 3D PSNR vs. $31.40\text{ dB}$ single-material NeAS baseline, a $+1.88\text{ dB}$ improvement). Furthermore, our model exhibits enhanced robustness under sparse-sampling conditions, outperforming baseline 3D PSNR by up to $1.17\text{ dB}$ under 5- and 10-view constraints.
\end{abstract}

\section{Introduction}
    Computed Tomography (CT) is widely used in clinical diagnosis but carries significant ionizing radiation risks~\cite{doi:10.1056/NEJMra072149}, which have been linked to increased lifetime cancer incidence, an effect that compounds with repeated CT visits. This motivates the development of sparse-view reconstruction methods that reduce patient dose while preserving diagnostic image quality. Recovering high-fidelity volumetric information from a limited number of projections is an ill-posed inverse problem, and classical analytical or iterative methods suffer from streak artifacts under sparse angular sampling.
    
    Implicit scene representations (ISR) have emerged as a promising paradigm for sparse-view CT reconstruction. Adapting NeRF~\cite{mildenhall2020nerf} to the Beer-Lambert law, methods such as~\cite{zha2022naf,DBLP:journals/corr/abs-2202-02171,shen2023nerpimplicitneuralrepresentation,9710599,nikolakakis2026ribpull} recover continuous attenuation fields from sparse projections without explicit voxel grids, with structure-aware priors~\cite{cai2024structureawaresparseviewxray3d} further improving fidelity under extreme sparsity. However, these methods lack explicit surface geometry, making tissue segmentation dependent on manual thresholding. Geometry-aware representations address this by jointly learning attenuation and surface geometry: Neural Attenuation Surfaces (NeAS)~\cite{zha2025neas} couples neural SDFs~\cite{wang2021neus} with attenuation networks for clean surface extraction, but relies on scene-specific attenuation bounds and a non-differentiable two-material selector, limiting scalability to real clinical data with multiple anatomical structures.

    We propose $K$-NeAS, which addresses these limitations through three contributions:
    \begin{itemize}
        \item a soft, fully differentiable sequential occupancy filter, built on a shared attenuation backbone with $K$ lightweight prediction heads, that resolves material membership as a pointwise function of local SDF occupancy for an arbitrary number of materials.
        \item an unsupervised Gaussian Mixture Model (GMM)~\cite{Reynolds2015} that automatically estimates per-material attenuation bounds by sampling the volume of a converged single-material prior, eliminating manual scene-specific tuning.
        \item a scheduled auxiliary floater regularization that suppresses spurious empty space geometry during the early stages of optimization.
    \end{itemize}
    Evaluated on four simulated Cone-Beam CT (CBCT) datasets~\cite{zha2022naf}, $K$-NeAS successfully scales to arbitrary material counts and improves 3D volumetric fidelity on abdomen, chest, and foot scenes, while empirical analysis indicates that GMM boundary miscalibration remains a limiting factor in high-contrast regions such as the Jaw.
    
\section{Methodology}
\subsection{Preliminaries}
    We build upon the NeAS framework~\cite{zha2025neas}, which jointly learns a neural SDF and attenuation field from sparse CT projections; we summarize its core components below.

    NeAS uses two neural networks: an SDF network $\Theta_{\text{sdf}}$ that outputs a signed distance value $d$ and latent feature vector, and an attenuation network $\Theta_{\text{att}}$ that takes that feature vector as input to predict the raw attenuation $\overline{\mu}$. The final attenuation coefficient is $\mu(\mathbf{x}) = \overline{\mu} \cdot \Omega(d,s)$, where $\Omega$ is the sigmoid-based surface boundary function (SBF) defined in Eq.~\ref{eq:SBF} and $s$ is a hyperparameter controlling the steepness of surface boundaries.
    \begin{equation}
        \Omega(d, s) = \frac{\exp(-sd)}{1+\exp(-sd)}
        \label{eq:SBF}
    \end{equation}
    
    When extended to two materials, $\Theta_{\text{sdf}}$ returns two signed distances $d_1, d_2$ and two disjoint attenuation MLPs $\Theta_{\text{att1}}$ and $\Theta_{\text{att2}}$ to predict raw attenuations $\overline{\mu}_1, \overline{\mu}_2$. The attenuation coefficients are calculated as $\mu_1(\mathbf{x}) = \overline{\mu}_1 \cdot \Omega(d_1,s)$ and $\mu_2(\mathbf{x}) = \overline{\mu}_2 \cdot \Omega(d_2,s)$. A hard selector $\Lambda$ resolves the final coefficient as $\mu(\mathbf{x}) = \Lambda(d_2, \mu_1, \mu_2)$, assigning $\mu_1$ when the point lies outside the inner surface $d_2 \geq 0$ and $\mu_2$ otherwise. This selector is non-differentiable at the boundary and does not generalize beyond two materials.  This model is optimized using an intensity MSE loss $\mathcal{L}_{\text{int}}$ and eikonal regularization~\cite{icml2020_2086} $\mathcal{L}_{\text{reg}}$ with material attenuation bounds $\alpha$ and $\beta$ requiring manual, scene-specific tuning. Additionally, NeAS employs frequency regularization~\cite{10.1007/978-3-031-72670-5_11} on the hash encoding to help regularize overfitting and poor camera pose estimation in sparse-view scenarios.

\subsection{$K$-NeAS}

\subsubsection{Architecture}
    While the disjoint networks of the baseline NeAS architecture isolate tissue representations, $K$-NeAS introduces a shared latent backbone designed to learn cross-material features, which is crucial for resolving ambiguous material boundaries for tissues with similar attenuation profiles. Each material $\Phi_i$, for $0 \le i \le K-1$, is given its own lightweight prediction head $\Theta_{\text{att}^i}$ branching from the shared backbone.
    
\begin{figure}[htbp]
        \centering
        \includegraphics[width=1\textwidth]{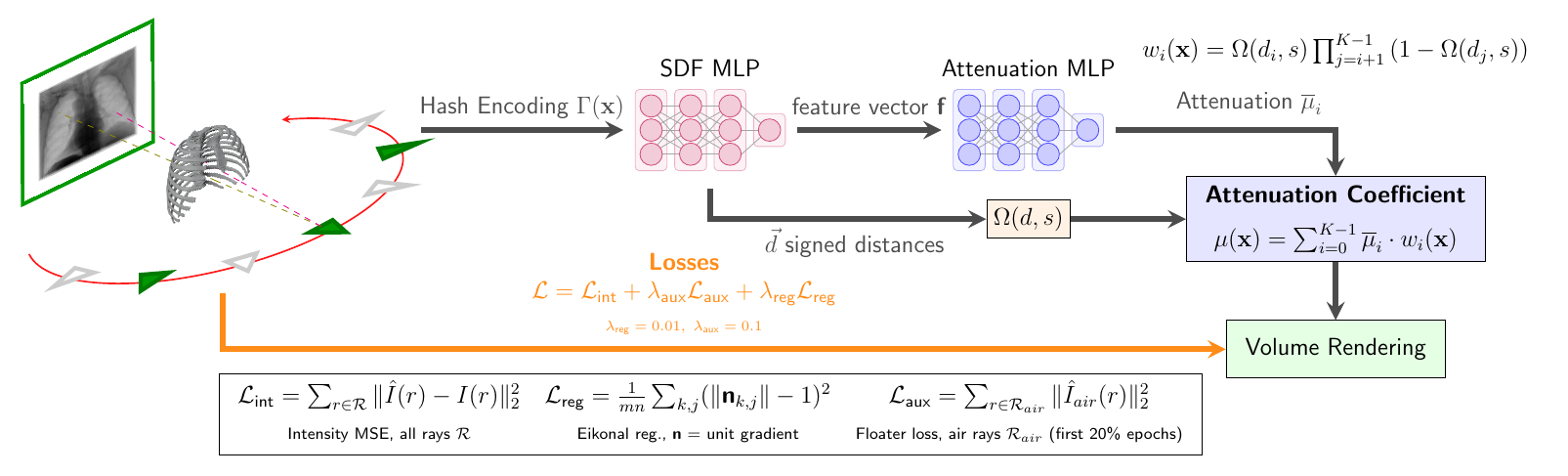}
        \caption{$K$-NeAS Model Architecture}
        \label{fig:proposed_arch}
    \end{figure}

    Figure \ref{fig:proposed_arch} illustrates the $K$-NeAS forward pass. A queried coordinate $\mathbf{x} \in \mathbb{R}^3$ is encoded via multi-resolution hash encoding $\Gamma(\mathbf{x})$ and passed through $\Theta_{\text{sdf}}$ to predict a shared feature $\mathbf{f}$ and signed distances $\vec{d}$ for each material, evaluated by the SBF $\Omega(d,s)$. In parallel, $\mathbf{f}$ is processed by the shared backbone $\Theta_{\text{att}^{body}}$ and distributed to $K$ heads $\Theta_{\text{att}^k}$, producing raw attenuations $\overline{\mu}_k$. These are aggregated via our differentiable soft selector (Section \ref{sec:selector}) to compute $\mu(\mathbf{x})$ for volume rendering.

\subsubsection{$K$-Material Soft Selector}
    \label{sec:selector}
\begin{figure}[htbp]
        \centering
        \includegraphics[width=0.8\textwidth]{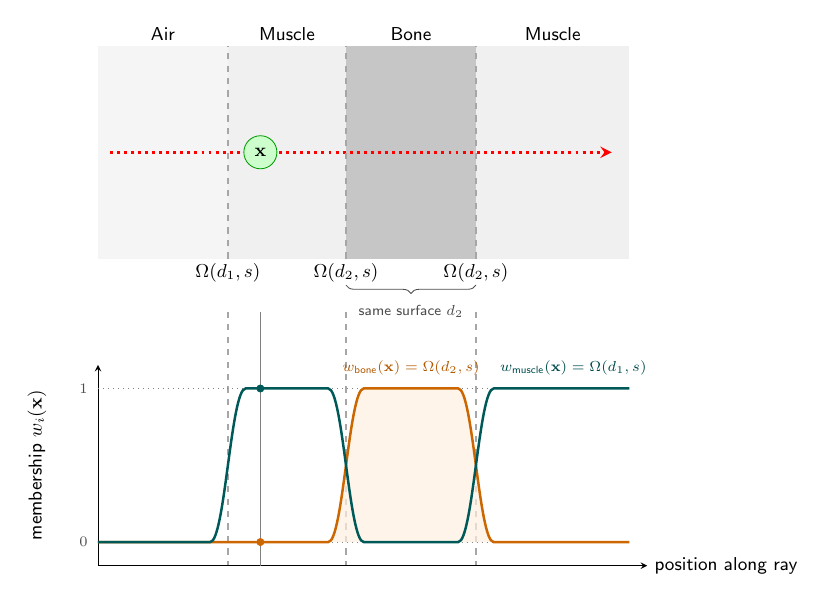}
        \caption{$K$-Selector, which assigns the membership of point $\mathbf{x}$ for two main materials: Muscle and Bone. The bottom diagram shows the weights $w_i(\mathbf{x})$ for each material.}
        \label{fig:selector}
    \end{figure}

    For scenes containing multiple anatomical structures, a spatial point $\mathbf{x}$ may lie within the overlapping surfaces of several materials. To resolve the ambiguity of the non-differentiable hard selector $\Lambda$ in two material NeAS, we introduce a soft, fully differentiable sequential occupancy filter (Figure~\ref{fig:selector}) that scales to arbitrary $K$ materials, addressing a limitation of NeAS. Inspired by the SDF-based compositional object rendering of ObjectSDF++~\cite{wu2023objectsdfplus}, our selector computes material membership as a pointwise function of local SDF occupancy at $\mathbf{x}$, rather than a quantity that accumulates transmittance along the ray or requires unconstrained volume fractions.
    
    Each material $\Phi_i$, for $i = 0, \dots, K-1$, corresponds to one output head of the shared attenuation backbone, with signed distance $d_i$ and raw attenuation $\overline{\mu}_i$. Head index is determined by the ascending-attenuation ordering produced by the GMM boundary estimation step (Section~\ref{sec:gmm}), such that $i=0$ corresponds to the lowest-attenuation material and $i=K-1$ to the highest-attenuation material. Rather than assigning each point $\mathbf{x}$ to a single material via a hard, non-differentiable rule, we assign each head a probability weight $w_i(\mathbf{x})$ reflecting the likelihood that $\mathbf{x}$ belongs to $\Phi_i$, discounted by the occupancy of higher-attenuation heads:
    \begin{equation}
        w_i(\mathbf{x}) = \Omega(d_i(\mathbf{x}), s) \prod^{K-1}_{j = i+1} \left(1 - \Omega(d_j(\mathbf{x}), s)\right)
        \label{eq:priority_weight}
    \end{equation}
    
    Intuitively, $\Omega(d_i(\mathbf{x}), s)$ gives the occupancy probability of $\Phi_i$, discounted by the occupancy of denser materials $\Phi_j$ with $j > i$. This simplifies to $w_{K-1}(\mathbf{x}) = \Omega(d_{K-1}, s)$ for the highest-attenuation material. The final attenuation coefficient is then the expected attenuation under this probability distribution:
    \begin{equation}
        \mu(\mathbf{x}) = \sum^{K-1}_{i=0} \overline{\mu}_i \cdot w_i(\mathbf{x})
        \label{eq:priority}
    \end{equation}
    This formulation remains fully differentiable and imposes no limit on material count. While real anatomical materials do not physically co-occupy the same spatial location, the soft weighting in Eq.~\ref{eq:priority} is designed to model partial-volume effects at tissue boundaries.
    
\subsubsection{Determining Attenuation Bounds}
    \label{sec:gmm}
    After training a single-material model with defaults $\alpha = 3.4, \beta = 0.1$ from NeAS~\cite{zha2025neas}, we fit a Gaussian Mixture Model (GMM)~\cite{Reynolds2015} to determine the optimal $\alpha$ and $\beta$ bounds for $K = 1, \dots, 4$ materials. Boundaries are placed at the density valley between adjacent Gaussian peaks, with the leftmost boundary fixed at zero and the rightmost set to 15\% above the 99.5th percentile of the distribution. For each interval $[a_i, b_i]$, we assign $\beta_i = a_i$ and $\alpha_i = b_i - a_i$ to ensure non-overlapping activation intervals across materials.
    
    \begin{figure}[htbp]
        \centering
        \includegraphics[width=0.8\textwidth]{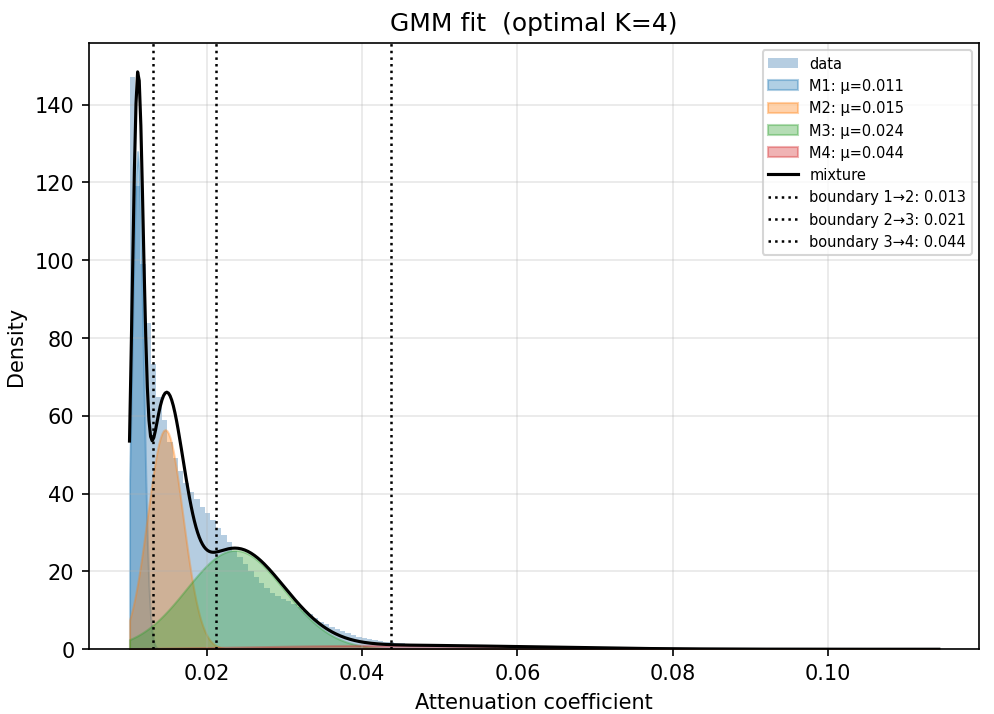}
        \caption{GMM fit (optimal $K=4$) to the Jaw attenuation density histogram, obtained from a preconverged single-material (1M) model.}        
        \label{fig:jaw_gmm}
    \end{figure}
    
\subsubsection{Floater Regularization}
    \label{sec:floater}
    Hash-encoded networks under sparse views tend to produce spurious geometry (``floaters'') in empty space. To suppress this, we decouple ray sampling into two independent batches: 512 rays from non-zero attenuation pixels for the primary loss $\mathcal{L}_{\text{int}}$, and 128 zero-attenuation rays for an auxiliary floater loss:
    \begin{equation}
        \mathcal{L}_{aux} = \sum_{r \in \mathcal{R}_{air}} \|\hat{I}_{air}(r)\|_2^2
        \label{eq:floater_loss}
    \end{equation}
    where $\hat{I}_{air}(r)$ is the predicted intensity along air rays, which should render to zero. $\mathcal{L}_{\text{aux}}$ is applied only during the first 20\% of training epochs, where floaters are most prevalent, preventing overfitting to zero-attenuation rays that could otherwise suppress valid low-attenuation features.
    
    The final loss to be optimized is as follows:
    \begin{equation}
        \mathcal{L} = \mathcal{L}_{\text{int}} + \lambda_{\text{aux}}\mathcal{L}_{\text{aux}} + \lambda_{\text{reg}}\mathcal{L}_{\text{reg}}
    \end{equation}
    We set $\lambda_{\text{aux}} = 0.1$ and $\lambda_{\text{reg}} = 0.01$ for all experiments.

\section{Experimental Setup}
    All models were trained on NVIDIA A4000 GPUs with 16GB of VRAM. Training runs for 1000 epochs per configuration, sampling 512 non-zero attenuation rays and 128 zero-attenuation rays, per projection per epoch, with the latter used exclusively during the first 20\% of training for floater regularization. The shared hash grid and SDF backbone account for $99.8\%$ of parameters ($\sim 12.167\times 10^6$); scaling $K=1\to4$ adds only $17{,}030$ parameters ($\le 0.14\%$ growth), keeping VRAM and compute roughly constant. The model was optimized using Adam~\cite{kingma2017adammethodstochasticoptimization}. We evaluate on the simulated CBCT dataset from NAF~\cite{zha2022naf}, covering four anatomical regions: Abdomen, Chest, Foot, and Jaw, each with 50 training and 50 held-out validation projections. As the NAF dataset provides calibrated camera poses, pose refinement and its associated frequency regularization are omitted from both baseline and proposed models. Because these simulated projections do not model real-acquisition effects such as beam hardening, scatter, or detector noise, and because ground-truth surface geometry is unavailable even for real clinical scans, evaluation is limited to image quality metrics on held-out projections. We test across four sparsity levels (5, 10, 20, 50 views), uniformly subsampled from the 50 training projections to ensure consistent angular coverage. We report PSNR and SSIM~\cite{1284395} in both the 2D projection-domain (on held-out validation predictions) and 3D volume-domain, providing a comprehensive measure of reconstruction fidelity.

\section{Results \& Analysis}
    \subsection{NeAS vs. $K$-NeAS}
\begin{figure}[htbp]
        \centering
        \caption{Comparisons between $K$-NeAS, ground truth, and NeAS. Abdomen, Chest, Foot, and Jaw are evaluated at 2 materials, using held-out validation projection indices 24, 1, 1, and 30, respectively. These are evaluated on a single run.}
        \includegraphics[width=0.9\textwidth]{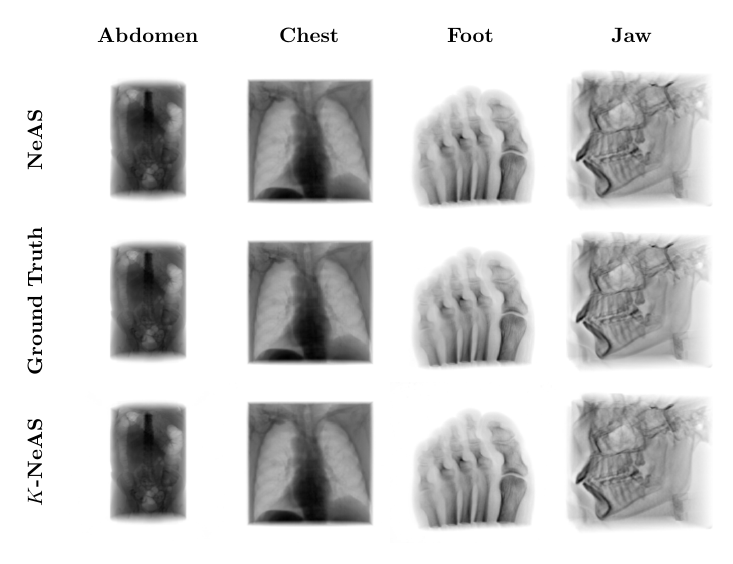}
        \label{fig:comparisons}
    \end{figure}

    Table~\ref{tab:results} presents quantitative comparisons between NeAS and our proposed pipeline across 2D and 3D PSNR and SSIM metrics. Qualitative projection comparisons are shown in Figure~\ref{fig:comparisons}.
    
    A primary limitation of NeAS is its restriction to two materials. As shown in Table~\ref{tab:results}, $K$-NeAS extends reconstruction to arbitrary material counts. In the Abdomen scene, fidelity scales positively with material complexity, peaking at 3M, indicating our soft selector effectively isolates distinct tissues. In Chest and Foot, $K$-NeAS matches or exceeds NeAS on 3D reconstruction metrics, with 3M achieving the best 3D PSNR in both; 2D projection-domain PSNR remains comparable to or slightly below the NeAS baseline in these scenes. Conversely, $K$-NeAS underperforms NeAS across all metrics in the Jaw scene, an edge case attributed to GMM boundary miscalibration (Section~\ref{sec:conclusion}). Notably, unlike NeAS, which requires an entirely separate attenuation network per material, $K$-NeAS's shared-backbone design adds only $17{,}030$ parameters when scaling from $K=1$ to $K=4$ ($\le 0.14\%$ growth, Section 3), so the flexibility to select an arbitrary material count comes at negligible additional training or inference cost.
    
    \begin{table}[htbp]
        \centering
        \caption{Quantitative comparison ($K$-NeAS results) across scenes and configurations. Best score per metric and scene in \textbf{bold}. Results are the average of 3 training runs.}
        \label{tab:results}
        \begin{tabular}{lccccccccc}
            \toprule
                & & \multicolumn{2}{c}{2D PSNR $\uparrow$} & \multicolumn{2}{c}{2D SSIM $\uparrow$} & \multicolumn{2}{c}{3D PSNR $\uparrow$} & \multicolumn{2}{c}{3D SSIM $\uparrow$} \\
            \cmidrule(lr){3-4} \cmidrule(lr){5-6} \cmidrule(lr){7-8} \cmidrule(lr){9-10}
                Scene & Config & NeAS & $K$-NeAS & NeAS & $K$-NeAS & NeAS & $K$-NeAS & NeAS & $K$-NeAS \\
            \midrule
            \multirow{4}{*}{Abdomen}
                 & 1M & 46.744 & 46.850 & 0.992 & 0.993 & 31.399 & 32.648 & 0.820 & 0.845 \\
                 & 2M & 47.098 & 47.204 & 0.993 & 0.993 & 31.391 & 32.722 & 0.823 & 0.846 \\
                 & 3M & ---    & \textbf{47.687} & ---   & \textbf{0.994} & ---    & \textbf{33.276} & ---   & \textbf{0.858} \\
                 & 4M & ---    & 47.346 & ---   & \textbf{0.994} & ---    & 32.916 & ---   & 0.850 \\
            \midrule
            \multirow{4}{*}{Chest}
                 & 1M & 45.949 & 45.230 & 0.991 & 0.991 & 31.508 & 31.396 & 0.913 & 0.908 \\
                 & 2M & \textbf{46.609} & 45.672 & 0.992 & 0.992 & 32.171 & 31.784 & 0.918 & 0.915 \\
                 & 3M & ---    & 46.036 & ---   & \textbf{0.993} & ---    & \textbf{32.181} & ---   & \textbf{0.920} \\
                 & 4M & ---    & 45.427 & ---   & 0.992 & ---    & 31.592 & ---   & 0.911 \\
            \midrule
            \multirow{4}{*}{Foot}
                 & 1M & 42.230 & 42.717 & 0.981 &\textbf{0.983}& 31.007 & 31.579 & \textbf{0.900} & 0.893 \\
                 & 2M & 42.224 & 42.570 & 0.981 & 0.982 & 31.092 & 31.388 & 0.891 & 0.889 \\
                 & 3M & ---    & \textbf{42.800} & ---   & \textbf{0.983} & ---    & \textbf{31.603} & ---   & 0.889 \\
                 & 4M & ---    & 42.717 & ---   & \textbf{0.983} & ---    & 31.462 & ---   & 0.887 \\
            \midrule
            \multirow{4}{*}{Jaw}
                 & 1M & \textbf{39.451} & 34.321 & \textbf{0.966} & 0.953 & \textbf{34.099} & 31.672 & \textbf{0.882} & 0.797 \\
                 & 2M & 37.472 & 34.366 & 0.963 & 0.956 & 33.514 & 31.828 & 0.832 & 0.808 \\
                 & 3M & ---    & 34.332 & ---   & 0.955 & ---    & 31.847 & ---   & 0.806 \\
                 & 4M & ---    & 34.541 & ---   & 0.956 & ---    & 31.879 & ---   & 0.810 \\
            \bottomrule
        \end{tabular}
    \end{table}
    
    While both models exhibit expected performance degradation as angular sampling decreases, the $K$-NeAS architecture demonstrates slightly higher robustness at moderate sparsity levels. Specifically, at the 10-view threshold, the proposed model outperforms the baseline in both 2D PSNR and 3D PSNR. At extreme sparsity (5 views), both models suffer significant degradation, which is reflected by sharp PSNR degradation, as shown in Table~\ref{tab:sparse_chest}.
    
    \begin{table}[htbp]
        \centering
        \caption{Chest Comparison across 5, 10, 20, and 50 views, on 2-material configurations for both NeAS and $K$-NeAS. }
        \label{tab:sparse_chest}
        \setlength{\tabcolsep}{4pt}
        \begin{tabular}{ccc|cc|cc|cc}
            \toprule
            & \multicolumn{2}{c}{2D PSNR $\uparrow$} & \multicolumn{2}{c}{2D SSIM $\uparrow$} & \multicolumn{2}{c}{3D PSNR $\uparrow$} & \multicolumn{2}{c}{3D SSIM $\uparrow$} \\
            \cmidrule(lr){2-3} \cmidrule(lr){4-5} \cmidrule(lr){6-7} \cmidrule(lr){8-9}
            \# Views & NeAS & $K$-NeAS & NeAS & $K$-NeAS & NeAS & $K$-NeAS & NeAS & $K$-NeAS \\
            \midrule
            5  & 30.473 & 30.813 & 0.899 & 0.913 & 20.596 & 21.767 & 0.502 & 0.554 \\
            10 & 35.058 & 35.997 & 0.940 & 0.948 & 23.414 & 24.486 & 0.638 & 0.673 \\
            20 & 40.307 & 40.788 & 0.972 & 0.973 & 26.424 & 27.112 & 0.760 & 0.775 \\
            50 & 46.609 & 45.672 & 0.992 & 0.992 & 32.171 & 31.784 & 0.918 & 0.915 \\
            \bottomrule
        \end{tabular}
    \end{table}
    
\subsection{Ablations}
    \begin{table}[t]
        \centering
        \caption{Component ablation on the Foot dataset (2-material configuration). Metrics reflect a single representative training run to isolate architectural contributions.}
        \setlength{\tabcolsep}{3pt}
        \label{tab:ablations}
        \begin{tabular}{lcccc}
            \toprule
            & 2D PSNR $\uparrow$ & 2D SSIM $\uparrow$ & 3D PSNR $\uparrow$ & 3D SSIM $\uparrow$ \\
            \midrule
            $K$-Selector                  & 41.938 & 0.991 & 31.071 & 0.884 \\
            $K$-Selector + Floater Reg    & 43.115 & 0.983 & 31.623 & 0.893 \\
            $K$-Selector + Shared Body    & 42.544 & 0.981 & 31.245 & 0.883 \\
            $K$-NeAS                      & 43.245 & 0.983 & 31.630 & 0.890 \\
            \bottomrule
        \end{tabular}
    \end{table}
    Table~\ref{tab:ablations} reports component ablation results on the Foot dataset, isolating the contribution of each proposed architectural module. The full $K$-NeAS pipeline achieves the highest 2D and 3D PSNR, demonstrating that the shared latent backbone, soft selector, and floater regularization act synergistically to maximize volumetric fidelity. Conversely, analyzing the multi-material progression on the Jaw dataset (Table~\ref{tab:results}) reveals a clear performance gap relative to the NeAS baseline. Because the Jaw scene consists primarily of high-contrast bone and air with minimal soft-tissue gradation, fitting an unsupervised GMM to the density histogram places inter-material boundaries suboptimally (Figure~\ref{fig:jaw_gmm}). This propagates miscalibrated activation intervals into all downstream configurations. Notably, increasing material complexity ($K=2 \to 4$) does not recover this deficit, indicating that GMM boundary sensitivity---rather than architectural capacity---dominates performance in high-contrast anatomical scenes.
    
\section{Conclusion \& Future Work}
    \label{sec:conclusion}
    We introduced $K$-NeAS, a scalable implicit reconstruction framework that overcomes the rigid two-material limitation of NeAS through a shared latent backbone, a differentiable sequential soft selector, and GMM-automated attenuation bounding. Evaluated across four simulated CBCT datasets, $K$-NeAS scales effectively to arbitrary anatomical complexity, achieving a $+1.88\text{ dB}$ 3D PSNR improvement over baseline on multi-tissue abdominal volumes at $K=3$ while maintaining superior robustness under sparse angular sampling.
    
    Despite these gains, two primary limitations highlight open challenges for future work. First, unsupervised GMM bound estimation remains vulnerable to sub-optimal valley detection in anatomically homogeneous scenes like the jaw, necessitating more constrained interval estimation. Furthermore, our evaluation relies on simulated CBCT projections and does not capture practical acquisition artifacts such as beam hardening, scatter, detector noise, or geometric miscalibration. Validating $K$-NeAS on real clinical acquisitions is an important direction for future work. As $K$-NeAS is designed as a direct architectural extension of NeAS, our evaluation focuses on a controlled comparison against this baseline under matched conditions; benchmarking against general-purpose CT reconstruction methods that do not model explicit surface geometry is left to future work. Second, preliminary experiments applying both NeAS and $K$-NeAS to cranial CT datasets resulted in severe degradation across both architectures, characterized by pervasive volumetric noise and surface boundary collapse. We suspect this instability stems from how our multi-resolution hash grid and coordinate-based ray sampling handle the extreme contrast between dense skull bone and soft tissue, introducing high-frequency noise artifacts along ray paths. Exploring alternative ray-sampling strategies or spatial encodings for such high-contrast volumes remains a promising direction for future research. Beyond addressing these limitations, we plan to expand our evaluation to additional anatomical regions, such as the brain and shoulder, to further validate the generalizability of $K$-NeAS across diverse clinical scenes.

    \paragraph{Code and Data Availability.} Code is publicly available at \url{https://github.com/dakshshah03/K-NeAS}. All data used are from the publicly available NAF CBCT dataset~\cite{zha2022naf}.

\begin{credits}
\subsubsection{\discintname}
The authors have no competing interests to declare that are relevant to the  content of this article.
\end{credits}

\bibliographystyle{splncs04}
\bibliography{references}

@misc{zha2025neas,
  title={NeAS: 3D Reconstruction from X-ray Images using Neural Attenuation Surface}, 
  author={Chengrui Zhu and Ryoichi Ishikawa and Masataka Kagesawa and Tomohisa Yuzawa and Toru Watsuji and Takeshi Oishi},
  year={2025},
  eprint={2503.07491},
  archivePrefix={arXiv},
  primaryClass={eess.IV},
  url={https://arxiv.org/abs/2503.07491}, 
}

@misc{wu2023objectsdfplus,
      title={ObjectSDF++: Improved Object-Compositional Neural Implicit Surfaces}, 
      author={Qianyi Wu and Kaisiyuan Wang and Kejie Li and Jianmin Zheng and Jianfei Cai},
      year={2023},
      eprint={2308.07868},
      archivePrefix={arXiv},
      primaryClass={cs.CV},
      url={https://arxiv.org/abs/2308.07868}, 
}

@misc{cai2024structureawaresparseviewxray3d,
      title={Structure-Aware Sparse-View X-ray 3D Reconstruction}, 
      author={Yuanhao Cai and Jiahao Wang and Alan Yuille and Zongwei Zhou and Angtian Wang},
      year={2024},
      eprint={2311.10959},
      archivePrefix={arXiv},
      primaryClass={eess.IV},
      url={https://arxiv.org/abs/2311.10959}, 
}

@ARTICLE{1284395,
  author={Zhou Wang and Bovik, A.C. and Sheikh, H.R. and Simoncelli, E.P.},
  journal={IEEE Transactions on Image Processing}, 
  title={Image quality assessment: from error visibility to structural similarity}, 
  year={2004},
  volume={13},
  number={4},
  pages={600-612},
  doi={10.1109/TIP.2003.819861}}

@misc{kingma2017adammethodstochasticoptimization,
      title={Adam: A Method for Stochastic Optimization}, 
      author={Diederik P. Kingma and Jimmy Ba},
      year={2017},
      eprint={1412.6980},
      archivePrefix={arXiv},
      primaryClass={cs.LG},
      url={https://arxiv.org/abs/1412.6980}, 
}

@InCollection{Reynolds2015,
    author      = {Douglas Reynolds},
    title       = {Gaussian Mixture Models},
    booktitle   = {Encyclopedia of Biometrics},
    doi         = {10.1007/978-1-4899-7488-4_196},
    pages       = {827--832},
    publisher   = {Springer {US}},
    year        = {2015}
}

@article{DBLP:journals/corr/abs-2202-02171,
  author       = {Darius R{\"{u}}ckert and
                  Yuanhao Wang and
                  Rui Li and
                  Ramzi Idoughi and
                  Wolfgang Heidrich},
  title        = {NeAT: Neural Adaptive Tomography},
  journal      = {CoRR},
  year         = {2022},
  url          = {https://arxiv.org/abs/2202.02171},
  eprinttype   = {arXiv},
  eprint       = {2202.02171},
  bibsource    = {dblp computer science bibliography, https://dblp.org}
}

@misc{shen2023nerpimplicitneuralrepresentation,
      title={NeRP: Implicit Neural Representation Learning with Prior Embedding for Sparsely Sampled Image Reconstruction}, 
      author={Liyue Shen and John Pauly and Lei Xing},
      year={2023},
      eprint={2108.10991},
      archivePrefix={arXiv},
      primaryClass={eess.IV},
      url={https://arxiv.org/abs/2108.10991}, 
}

@INPROCEEDINGS{9710599,
  author={Zang, Guangming and Idoughi, Ramzi and Li, Rui and Wonka, Peter and Heidrich, Wolfgang},
  booktitle={2021 IEEE/CVF International Conference on Computer Vision (ICCV)}, 
  title={IntraTomo: Self-supervised Learning-based Tomography via Sinogram Synthesis and Prediction}, 
  year={2021},
  volume={},
  number={},
  pages={1940-1950},
  doi={10.1109/ICCV48922.2021.00197}}

@article{doi:10.1056/NEJMra072149,
author = {David J. Brenner  and Eric J. Hall },
title = {Computed Tomography — An Increasing Source of Radiation Exposure},
journal = {New England Journal of Medicine},
volume = {357},
number = {22},
pages = {2277-2284},
year = {2007},
doi = {10.1056/NEJMra072149},

URL = {https://www.nejm.org/doi/full/10.1056/NEJMra072149},
eprint = {https://www.nejm.org/doi/pdf/10.1056/NEJMra072149}
}

@inproceedings{zha2022naf,
  author    = {Zha, Ruyi and Zhang, Yanhao and Li, Hongdong},
  title     = {NAF: Neural Attenuation Fields for Sparse-View CBCT Reconstruction},
  booktitle = {International Conference on Medical Image Computing and Computer-Assisted Intervention (MICCAI)},
  pages     = {442--452},
  year      = {2022},
  organization={Springer}
}

@inproceedings{mildenhall2020nerf,
 title={NeRF: Representing Scenes as Neural Radiance Fields for View Synthesis},
 author={Ben Mildenhall and Pratul P. Srinivasan and Matthew Tancik and Jonathan T. Barron and Ravi Ramamoorthi and Ren Ng},
 year={2020},
 booktitle={ECCV},
}

@incollection{icml2020_2086,
 author = {Gropp, Amos and Yariv, Lior and Haim, Niv and Atzmon, Matan and Lipman, Yaron},
 booktitle = {Proceedings of Machine Learning and Systems 2020},
 pages = {3569--3579},
 title = {Implicit Geometric Regularization for Learning Shapes},
 year = {2020}
}

@article{wang2021neus,
  title={NeuS: Learning Neural Implicit Surfaces by Volume Rendering for Multi-view Reconstruction},
  author={Wang, Peng and Liu, Lingjie and Liu, Yuan and Theobalt, Christian and Komura, Taku and Wang, Wenping},
  journal={arXiv preprint arXiv:2106.10689},
  year={2021}
}

@inproceedings{10.1007/978-3-031-72670-5_11,
author = {Xie, Shuxiang and Zhou, Shuyi and Sakurada, Ken and Ishikawa, Ryoichi and Onishi, Masaki and Oishi, Takeshi},
title = {G2fR: Frequency Regularization in Grid-Based Feature Encoding Neural Radiance Fields},
year = {2024},
isbn = {978-3-031-72669-9},
publisher = {Springer-Verlag},
address = {Berlin, Heidelberg},
url = {https://doi.org/10.1007/978-3-031-72670-5_11},
doi = {10.1007/978-3-031-72670-5_11},

booktitle = {Computer Vision – ECCV 2024: 18th European Conference, Milan, Italy, September 29 – October 4, 2024, Proceedings,  Part XXII},
pages = {186–203},
numpages = {18},
location = {Milan, Italy}
}

@inproceedings{nikolakakis2026ribpull,
  title={{RibPull}: implicit occupancy fields and medial axis extraction for {CT} ribcage scans},
  author={Nikolakakis, Emmanouil and Ouasfi, Amine and Digne, Julie and Marinescu, Razvan},
  booktitle={Medical Imaging 2026: Image Processing},
  volume={13925},
  pages={414--420},
  year={2026},
  organization={SPIE},
  doi={10.1117/12.3085604},
  url={https://www.spiedigitallibrary.org/conference-proceedings-of-spie/13925/139251I/RibPull--implicit-occupancy-fields-and-medial-axis-extraction-for/10.1117/12.3085604.short}
}

\end{document}